\documentclass[letterpaper]{article} 
\usepackage{aaai2026}  
\usepackage{times}  
\usepackage{helvet}  
\usepackage{courier}  
\usepackage[hyphens]{url}  
\usepackage{graphicx} 
\usepackage{natbib}  
\usepackage{caption} 
\usepackage{algorithm}
\usepackage{algorithmic}

\usepackage{newfloat}
\usepackage{listings}
\DeclareCaptionStyle{ruled}{labelfont=normalfont,labelsep=colon,strut=off} 
\floatstyle{ruled}
\newfloat{listing}{tb}{lst}{}
\floatname{listing}{Listing}
\usepackage{latexsym}
\usepackage{amssymb}
\usepackage{amsmath}
\usepackage{amsthm}
\usepackage{booktabs}
\usepackage{enumitem}
\usepackage{multirow}
\usepackage{tabularx}
\usepackage{makecell}
\usepackage{colortbl}
\usepackage[table,xcdraw]{xcolor}

\newcommand{\OurFed}{LSHFed}
\newcommand{\LSH}{LSHGM}

\newcolumntype{Y}{>{\centering\arraybackslash}X}       
\newcolumntype{M}[1]{>{\raggedright\arraybackslash}m{#1}} 

\title{LSHFed: Robust and Communication-Efficient Federated Learning with Locally-Sensitive Hashing Gradient Mapping}
\author{
    Guanjie Cheng\textsuperscript{\rm 1},
    Mengzhen Yang\textsuperscript{\rm 1},
    Xinkui Zhao\textsuperscript{\rm 1},
    Shuyi Yu\textsuperscript{\rm 1},
    Tianyu Du\textsuperscript{\rm 1},
    Yangyang Wu\textsuperscript{\rm 1},
    Mengying Zhu\textsuperscript{\rm 1},
    Shuiguang Deng\textsuperscript{\rm 2}
}
\affiliations{
    \textsuperscript{\rm 1}School of Software Technology, Zhejiang University\\
    \textsuperscript{\rm 2}College of Computer Science and Technology, Zhejiang University\\
    \{chengguanjie,mzyang1610,zhaoxinkui,yushuyi\}@zju.edu.cn,tydusky@gmail.com,\{zjuwuyy,mengyingzhu,dengsg\}@zju.edu.cn
}

\begin{document}

\maketitle

\begin{abstract}
Federated learning (FL) enables collaborative model training across distributed nodes without exposing raw data, but its decentralized nature makes it vulnerable in trust-deficient environments. Inference attacks may recover sensitive information from gradient updates, while poisoning attacks can degrade model performance or induce malicious behaviors. Existing defenses often suffer from high communication and computation costs, or limited detection precision. To address these issues, we propose LSHFed, a robust and communication-efficient FL framework that simultaneously enhances aggregation robustness and privacy preservation. At its core, LSHFed incorporates LSHGM, a novel gradient verification mechanism that projects high-dimensional gradients into compact binary representations via multi-hyperplane locally-sensitive hashing. This enables accurate detection and filtering of malicious gradients using only their irreversible hash forms, thus mitigating privacy leakage risks and substantially reducing transmission overhead. Extensive experiments demonstrate that LSHFed maintains high model performance even when up to 50\% of participants are collusive adversaries while achieving up to a \textbf{1000×} reduction in gradient verification communication compared to full-gradient methods.
\end{abstract}


\section{Introduction}

Federated learning (FL) features collaboratively training models across distributed nodes. It allows different nodes to train on distinct datasets, enabling users who are reluctant to disclose private data to participate in training without the need to publicly share or transfer their datasets. 

FL has gained significant attention in both academia and industry \cite{han2023practical,jiang2023fair}. This growing popularity has spurred the adoption of FL in innovative solutions such as LLMs and fine-tuning methods. For example, Kuang et al. \cite{kuang2024federatedscope} leveraged FL to collaboratively enhance pre-trained LLMs for specialized applications, all while ensuring data privacy and reducing resource consumption.


However, the distributed architecture of FL renders it vulnerable to various challenges in trust‐deficient environments. Several works indicate that the advancement of FL is predominantly impeded by three interrelated challenges: \textit{persistent privacy-and-security vulnerabilities, pronounced heterogeneity in client-side data and prohibitive communicational overhead} \cite{cui2021malicious,liu2022distributed,wen2023survey}. Among the first challenge, two types of threats stand out as particularly critical for FL systems:

\textbf{(1) Poisoning attacks:} Malicious nodes can disrupt the aggregation process by uploading modified or tailored gradients \cite{liu2024survey}, resulting in performance degradation in the aggregated model or inducing attacker-specified behaviors for specific tasks.

\textbf{(2) Inference attacks:} Attackers aim to extract private training data from other participants \cite{liu2024survey}. Attackers may infer whether specific content exists in training sets—or even reconstruct partial training datasets—by analyzing gradient updates during the aggregation process.

Numerous methods have been proposed to defend against these attacks, including robust aggregation (e.g., Geometric Median \cite{pillutla2022robust}, Marginal Trimmed Mean \cite{nishimoto2023fedatm}), selection-based filtering (Krum, Multi-Krum \cite{colosimo2023median}), and clustering or similarity-based detection using Euclidean or cosine metrics \cite{stallmann2022towards,geiping2020inverting,nguyen2022flame,cao2020fltrust,awan2021contra}. While effective, many of these schemes incur substantial computational overhead. On the privacy side, techniques such as homomorphic encryption \cite{aziz2023exploring}, secure MPC \cite{liu2024survey}, and differential privacy \cite{xiong2021privacy,badar2024trustfed} have been employed to safeguard gradient exchanges. Concurrently, communication-efficient strategies—including sparse matrix representations \cite{li2023losparse} and gradient quantization \cite{mai2024rflpa,xu2021signguard,peng2023birder,oh2022communication}—reduce transmission costs but may degrade the accuracy of malicious update identification.


To address the aforementioned challenges, we propose \OurFed—a novel robust FL framework characterized by poisoning resistance and privacy preservation. Specifically, our Locally-Sensitive Hashing Gradient Mapping (\LSH) algorithm precisely screens for malicious gradients while reducing communication overhead to 0.07\% of that required to transmit full gradients. A distributed masking strategy is designed to achieve end-to-end privacy protection.

The main contributions of this work are:

(1) We develop a communication-efficient and maliciousness-sensitive gradient detection algorithm, namely \LSH, which enables malicious gradient identification while only transmitting 0.07\% of the original gradient at minimum, ensuring both computation and communication efficiency.

(2) Building upon the \LSH~algorithm, we integrate a mask-based privacy protection mechanism and a node reputation mechanism, namely the ScoreQ-Hash election algorithm, to establish \OurFed—a robust and privacy-preserving FL framework.

(3) Comprehensive experimental evaluations verify that \OurFed~remains resilient to malicious attacks in FL tasks and consistently surpasses representative baseline methods across security and utility metrics.

\section{Related Work}

\subsection{Poisoning Attacks}
To defend against poisoning, robust aggregation rules and filtering strategies have been developed. Geometric Median \cite{wang2023wgm,pillutla2022robust} and Marginal Trimmed Mean \cite{nishimoto2023fedatm} mitigate outliers by computing central statistics or discarding extremes. Krum and its variants select updates closest to the majority \cite{bouhata2024byzantine,colosimo2023median}, though their complexity and sensitivity to non‐IID data can hinder scalability. Clustering and similarity‐based methods further filter malicious updates via Euclidean distance \cite{herath2023recursive} or cosine similarity \cite{ZHU2024110730,stallmann2022towards}, as exemplified by ShieldFL \cite{ma2022shieldfl}, FLAME \cite{nguyen2022flame}, FLTrust \cite{cao2020fltrust}, and Contra \cite{awan2021contra}.

\subsection{Privacy Leakages}
Privacy‐preserving mechanisms in FL include homomorphic encryption \cite{aziz2023exploring}, secure MPC \cite{liu2024survey}, and differential privacy \cite{emelianov2022fair,xiong2021privacy}. Local DP offers finer granularity but may degrade utility. RFLPA \cite{mai2024rflpa} integrates verifiable Shamir secret sharing with a dot‐product aggregation to balance privacy and malicious detection, while TrustFed \cite{badar2024trustfed} employs Gaussian noise masking under LDP. Techniques such as noise‐direction alignment \cite{wu2025privacy} and noise‐summation correction (VerifyNet) \cite{xu2019verifynet} seek to mitigate accuracy loss.

\subsection{Communication \& Computation Optimization}
Efficient aggregation leverages sparsification \cite{wang2020adaptive}, sign‐based compression \cite{xu2021signguard,seide20141}, and quantization \cite{oh2022communication}. While these reduce communication, they can impair detection accuracy or impose computational costs. RFLPA \cite{mai2024rflpa} further enhances overhead reduction compared to BREA \cite{so2020byzantine} by combining secret sharing with compressed updates.

\subsection{Locally‐Sensitive Hashing}
LSH enables sublinear nearest‐neighbor search in high dimensions \cite{yuan2023recent,jafari2021survey}. Recent adaptations include cross‐polytope hashing for MoE token clustering \cite{nie2024lsh}, reducing all‐to‐all communication with error compensation, and K‐HashFed’s integration of K‐means and LSH to compress and merge similar gradient centroids \cite{kapoor2025k}, achieving massive transmission savings under non‐IID settings.

\section{Method}

\subsection{Threat Model \& Design Overview}

\begin{figure*}[htbp]
\centering
\includegraphics[width=0.96\textwidth]{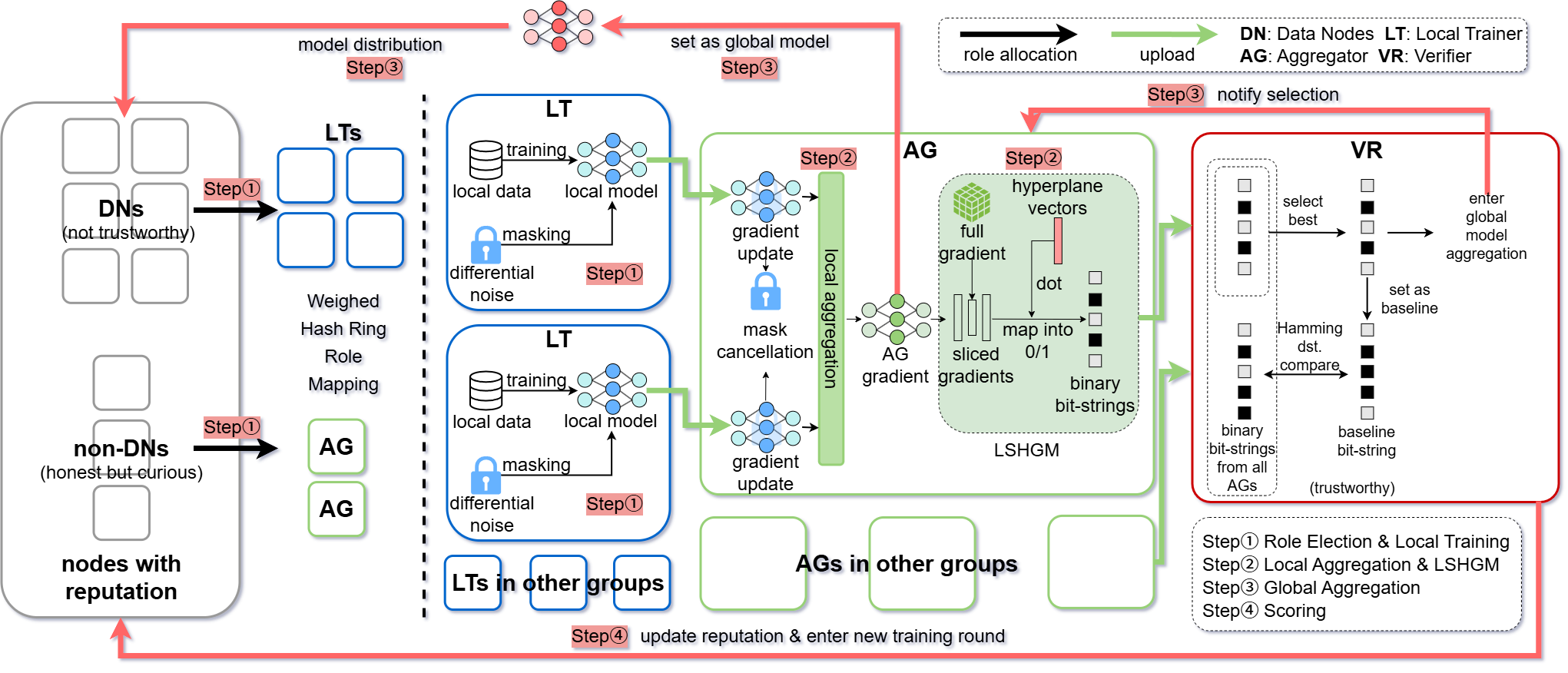}
\caption{\OurFed~Framework}
\label{framework}
\end{figure*}

\OurFed~framework and its general working process are shown in Figure \ref{framework}. \OurFed~divides participants into three roles: Verifier (VR), Aggregators (AG), and Local Trainers (LT). LTs are selected from nodes that have datasets and are willing to train the global model (referred to as DataNodes, DN). AGs are chosen from nodes without datasets (non-DN). We assume that VRs are trustworthy participants such as rule makers, regulatory authorities and the task publishers. Since VR carries little computation and communication overhead, it can be deployed at a low cost. Furthermore, non-DNs will report data honestly but are curious about privacy data independently. DNs are not trustworthy, meaning that DNs may conduct poisoning attack or upload fake data.

Each AG is in charge of the aggregation of a group of LTs, and the VR is in charge of all the nodes, including role election. LTs train models on their local datasets and upload gradient updates. AGs aggregate the gradient updates uploaded by LTs within their group to form a group-level gradient update and submit its \LSH~result to the VR. The VR then verifies and selects the best one to aggregate into the global model, and notify the corresponding AG to perform the aggregation and distribute the aggregated model to other active nodes for the next round of training. When the entire training process is finished, VR collects the final global model from the selected AG of the previous round.

Throughout the training process, the \OurFed~~framework can be divided into three main components: \LSH~Malicious Gradient Detection, the Mask-Based Privacy Protection Scheme, and the ScoreQ-Hash Role Election. The following sections will introduce these three components in order, followed by a detailed illustration of the compleSte training workflow.

\subsection{\LSH~Malicious Gradient Detection}
\LSH~is a multi‐hyperplane projection and bit string encoding algorithm that significantly reduces communication and storage overhead while preserving essential gradient features for effective malicious behavior recognition. It irreversibly projects gradients onto computable binary bit strings for malicious gradient detection. The detailed mechanism is as follows: 

\textbf{Generate Hyperplane Vectors. }
Suppose the gradient update of the current FL model is represented by \( N \) matrices \(\{M_1, M_2, \dots, M_N\}\). The algorithm generates \( r \) certain hyperplane vectors \( H_{i,k} ,i \in [1,N], k \in [1,r]\) for each gradient matrix.

\textbf{Column Segmentation of the Gradient Matrix. }
The algorithm segments each gradient matrix \( M_i ,i \in [1,N]\) into multiple column vectors. This block-wise processing captures gradient features by allowing each column vector to independently reflect its characteristics in the low-dimensional representation.

\textbf{Generate Bit String Results. }
The algorithm computes the dot product between each column vector and its corresponding hyperplane vector to generate a binary result. Assume that matrix \( M_i \) has \(m_i\) rows and \(n_i\) columns, for each segmented column vector \( C_{i,j} ,j \in [1,n_i]\) and hyperplane vector \( H_{i,k} ,k \in [1,r]\), the dot product is computed as:

\begin{eqnarray}\label{eq:vcg}
s_{i,j,k} = H_{i,k} \cdot C_{i,j}
\end{eqnarray}
By applying the sign function, the dot product result \( s_{i,j,k} \) is converted into a binary encoding value:

\begin{eqnarray}\label{eq:vcg}
f(s_{i,j,k}) = 
\begin{cases}
1, & \text{if } s_{i,j,k} \geq 0 \\
0, & \text{if } s_{i,j,k} < 0
\end{cases}
\end{eqnarray}

In this way, each column vector \( C_{i,j} \) is mapped to a binary value under the hyperplane vector \( H_{i,k} \), namely the generated bit strings. Bit string mapping calculation is irreversible, thus the exposure of hyperplane vectors \( H_{i,k} ,i \in [1,N], k \in [1,r]\) is privacy-secure.

\textbf{Bit String Comparison and Anomaly Detection. } 
By comparing the bit string generated by each AG with the benchmark bit string from the previous round, the algorithm computes their similarity to identify potential malicious behavior. Based on the gradient update result from the previous round, a benchmark bit string \(B = [B_1, B_2, \dots, B_{n}]\) is generated using the same hyperplane vectors. In the first round of the entire training process, the \LSH~benchmark bit string needs to be obtained through local training by a trusted entity: in our experimental setup, this is performed by the VR. 
Then, compute the Hamming distance between the current node's generated bit string \(
B' = [B'_1, B'_2, \dots, B'_{n}]
\)and the benchmark bit string \( B \), which can be efficiently calculated using a XOR operation. For each node, the Hamming distance between its generated bit string and the benchmark is computed and then sorted. \LSH~considers the gradient updates of nodes with a Hamming distance significantly higher than that of other nodes as malicious—in our setup, it only adopts gradients with the lowest Hamming distances.

\subsection{Mask-Based Privacy Protection Scheme}
Before each round of training, LTs with the same group generate their privacy masks based on a total mask sum \( R_{\text{sum}} \). This total sum is a VR-defined constant that all nodes know. Each node’s mask value is computed based on this total sum and the mask values of its preceding nodes. During aggregation, AGs subtract \(R_{\text{sum}}\) to remove all added noise.

In each training round, node \( n_j \) generates a random mask matrix \( M_{j,i} \) and calculates a residual value \( r_i \) that needs to be passed on. Specifically, node \( n_i \), based on the known sum of the masks of its preceding nodes, computes the residual value \( r_i \) to be passed to the next node using:
\begin{eqnarray}\label{eq:vcg}
r_i = R_{\text{sum}} - \sum_{k=1}^{i-1} M_{j,k} \mod d
\label{eq:remaining_value}
\end{eqnarray}
where \(\sum_{k=1}^{i-1} M_{j,k}\) represents the total mask generated by the first \(i-1\) nodes, and \( d \) is the largest mask value. Node \( n_i \) uses this residual \( r_i^t \) to compute its own mask and then passes the residual value to the next node. Each node only computes the residual based on the known preceding masks without having to disclose its own generated mask.

Finally, after the last node receives the residual values from all preceding nodes, the correctness of the mask generation can be verified using:
\begin{eqnarray}\label{eq:vcg}
\sum_{i=1}^n M_{j,i} = R_{\text{sum}} \quad \mod \; d
\label{eq:mask_verification}
\end{eqnarray}

\subsection{ScoreQ-Hash Role Election}

ScoreQ-Hash is a reputation-driven, weighted hash-ring election used to assign LT and AG roles dynamically. In the first round, \(P\) AGs are chosen uniformly at random from non-DNs, and all DNs act as LTs (suppose we have $R_{non-DN}$ non-DNs, then $P \leq R_{non-DN}$). From the second round onward, each node \(n_i\) receives a reputation score:
\begin{eqnarray}
S(i)=\alpha_1\,\phi^{(1)}_Q(i)\;+\;\alpha_2\,\phi^{(2)}_Q(i)\,
\end{eqnarray}
where \(\alpha_1\) and \(\alpha_2\) are reputation weights, satisfying \(\alpha_1+\alpha_2=1\). \(\phi^{(1)}_Q\) and \(\phi^{(2)}_Q\) are ranked time-consumption and \LSH-distance scores. \(\phi^{(1)}_Q\) is ranked from shortest time to longest, and \(\phi^{(2)}_Q\) is ranked from smallest \LSH-distance to longest. Both scores are mapped into range $[0,1]$ using quantile normalization. The quantile score is calculated using \(\phi^{(n)}_{\text{Q}}(i) = \frac{R - r^{(n)}_i}{R - 1}\), where  $r^{(n)}_i$ is the ranking of node $i$ in score $\phi^{(n)}$, $R$ is the total node count. Since the reputation score is memoryless, malicious nodes that initially behave honestly will still have their selection probability promptly reduced once misbehavior is detected.

DNs and non-DNs are then mapped onto two separate hash rings: each node’s arc length is proportional to its normalized weight:
\begin{eqnarray}
P(i)=\frac{S(i)}{\sum_j S(j)}\,
\end{eqnarray}
A fresh random start \(H_0=\mathsf{Hash}(\mathit{round\_id}\,\|\,\mathit{seed})\) determines the traversal point. Moving clockwise from \(H_0\), the next \(Q\) DNs become LTs and the next \(P\) non-DNs become AGs (suppose we have $R_{DN}$ DNs, then $Q \leq R_{DN}$). This mechanism ensures that higher-scoring nodes occupy larger ranges on the ring—and thus have a greater probability of selection—while low-score or malicious nodes, whose reputation falls with each detection event, are progressively excluded.

\subsection{General Workflow}

In this section, we illustrate the workflow of a training round in \OurFed:

\textbf{Step 1: Role Election and Local Training. }
Before training begins, VR selects active DNs as LTs and non-DNs as AGs , which is decided using the ScoreQ-Hash role election algorithm. Meanwhile, LT groups are decided by order. LTs within the same group collaboratively generate encryption masks for gradients distributively. Subsequently, each LT computes gradient updates using its local dataset. After generating gradients, each LT applies a privacy-preserving mask \( M_{j,i} \) on the original gradient \( g_{j,i} \) as \( \tilde{g}_{j,i} = g_{j,i} + M_{j,i} \).The masked gradient is then transmitted to the designated aggregation node \( AG_j^{t} \) within the group.  

\textbf{Step 2: Local Aggregation and \LSH~Bit String Generation. }
The AG node \( AG_j^{t} \) collects privacy-masked gradient updates from \( k \) LT nodes in the group and computes the locally aggregated gradient. Using the \LSH~algorithm, \( AG_j^{t} \) generates a compact bit string representation \( B_j^{t} \).

\textbf{Step 3: Global Aggregation. }
Upon receiving all bit strings from AG nodes, the VR computes the Hamming distance \( d_j^{t} \) between each bit string \( B_j^{t} \) and the previous round’s chosen bit string \( B^{t-1} \). VR ranks aggregation results by sorting Hamming distances in ascending order and select candidates within a predefined threshold as valid aggregation results. VR then notifies the selected AG to update the global model and distributes the updated model to other active nodes. 

\textbf{Step 4: Scoring. }
After completing a round of aggregation, VR scores nodes for the election in the next round. 

\subsection{Security Analysis}

In this section, we analyze \OurFed’s resilience against both privacy‐and integrity‐oriented threats under our assumed threat model. We show that (1) honest‐majority mask generation and differential‐privacy masking guarantee honest nodes’ gradient confidentiality, (2) irreversible \LSH~ensures robust detection of malicious gradients, and (3) ScoreQ‑Hash election prevents adversarial takeover.

\textbf{(1) Gradient Privacy.}  
Formally, let \(\tilde g_{j,i} = g_{j,i} + M_{j,i}\) be the masked gradient; then for any coalition of up to \(k < n\) LTs (with \(n\) the group size), the joint distribution of their masks is uniform over all vectors summing to at most \(R_{\mathrm{sum}}\), so the masked outputs \(\{\tilde g_{j,i}\}\) remain statistically independent of the true gradients \(\{g_{j,i}\}\). This achieves information‐theoretic privacy of individual updates under the honest‐majority assumption.

\textbf{(2) Irreversibility of \LSH.}  
In \LSH—where each gradient matrix is projected onto given hyperplanes and quantized to a bit string—no adversary (even a curious AG) can invert the bit string to recover information about the underlying gradient. Thus, our scheme provably bounds per‐round privacy loss while providing provable non‐invertibility of any transmitted summary.

\textbf{(3) Robustness.}  
Under the assumption that malicious LTs can control at most a fraction \(\alpha < \frac{1}{2}\) of participants, classical results on Hamming‐distance–based outlier detection guarantee that the honest updates cluster around the previous‐round benchmark bit string, while malicious outliers incur significantly larger distances. Sorting distances and selecting the lowest‐\((1-\alpha)\) fraction provably excludes all adversarial updates whenever the distance gap between honest and malicious clusters exceeds a threshold \(\tau\); in practice, we observe this gap to be large under common poisoning norms. Hence, \OurFed’s global aggregation resists up to \(\alpha\) Byzantine faults with high probability.

\textbf{(4) Secure Role Election.}  
ScoreQ‑Hash assigns selection weights proportional to a combination of past reliability and \LSH~ranks. Since the hash‐ring injects fresh randomness each round (\(H_0=\mathsf{Hash}(\mathit{round\_id}\,\|\,\mathit{seed})\)), no coalition of compromised non‑DNs can bias the choice of VR beyond their proportional reputation share. As long as a majority of high‐reputation nodes remain honest, the probability that an attacker controls all AG groups in any round is negligible.

\textbf{(5) Mask‑Integrity Robustness.}  
A malicious LT that uploads incorrect privacy mask introduces an irrecoverable residual noise \(\delta \;=\;\sum_{i\in \mathcal{G}} m_{j,i}^t \;-\; R_{\mathrm{sum}}\;\neq\;0\) into its group’s aggregated gradient. Since AGs subtract only \(R_{\mathrm{sum}}\), the noisy aggregate becomes \(\tilde G + \delta\). Under \LSH, this mask‑induced perturbation is the same as noise injected by an attacker, yielding a bit string with a Hamming distance greater than that of honest gradients and thus filtered out as an outlier.


Consequently, the corresponding node’s gradient score \(
\phi^{(2)}_Q(i)\;=\;\frac{N - r^{(2)}_i}{N-1}
\) collapses to its minimum quantile (where \(r^{(2)}_i\) is its rank among \(N\) nodes), and its overall reputation drops. This in turn reduces its selection weight.
Because ScoreQ‑Hash re‑elects roles with fresh randomness each round, the malicious node cannot hide in a stable honest group. Repeated mask tampering thus drives \(P(i)\to 0\), making the attacker increasingly unlikely to be chosen as an LT and effectively excluding it from future rounds.

Combining the above guarantees, \OurFed~achieves:
\begin{itemize}
\item \emph{Confidentiality}: No individual gradient is revealed to any other party beyond what is exposed by the final, noise‑free global model.  
\item \emph{Robustness}: Malicious gradient injections or model‐poisoning attempts are detected and excluded in every round, ensuring convergence to a clean model.  
\item \emph{Liveness}: Honest nodes are guaranteed eventual participation in both LT and AG roles, so progress is not stalled by adversarial denial‑of‑service.  
\end{itemize}

\section{Experiments}

\subsection{Experimental Setup}
We assess the performance and robustness of \OurFed~in image classification tasks, evaluating \OurFed~with five baseline FL frameworks on three public datasets. The detailed experiment configuration is shown in \ref{tab:exp-config}. The MNIST dataset is trained using a CNN model with two convolution layers, one pooling layer and two full connect layers, and CIFAR-10 is trained using a ResNet model, identical to the setup of RFLPA \cite{mai2024rflpa}. We also evaluate \OurFed~on the F-MNIST dataset using a vision transformer (ViT) and non-iid setups to fully assess its performance and scalability.

The experiments are conducted on a 24‑core Ubuntu 20.04 server with an RTX4090, 128GB RAM, and a 20‑core Windows laptop with an RTX4070, 32GB RAM. All code is implemented in Python 3.9.


\begin{table}[h]
  \centering
  \resizebox{0.48\textwidth}{!}{%
  \begin{tabular}{l|ccc}
    \toprule
    \textbf{Param} & \textbf{MNIST} & \textbf{CIFAR} & \textbf{FMNIST} \\
    \midrule
    \textbf{Number of Classes} & 10 & 10 & 10 \\
    \textbf{Model Type} & CNN & ResNet & ViT \\
    \textbf{Parameter Amount} & 206{,}922 & 6{,}575{,}370 & 805{,}130 \\
    \textbf{Model Structure} & c-c-p-f-f & c-c-r-c-c-r & e-n-h \\
    \textbf{DN Count (Total/LT)} & 10 / 5 & 10 / 5 & 10 / 5 \\
    \textbf{non-DN Count (Total/AG/VR)} & 10 / 2 / 1 & 10 / 2 / 1 & 10 / 2 / 1 \\
    \textbf{LSHGM Filter Rank} & 1 & 1 & 1 \\
    \bottomrule
  \end{tabular}%
  }
  \caption{Experiment Configuration.}
  \label{tab:exp-config}
  \footnotesize{\textit{*Model Structure: \textbf{c}=Convolution, \textbf{f}=Fully Connected, \textbf{r}=ResNet, \textbf{p}=Pooling, \textbf{e}=Transformer Encoder, \textbf{n}=LayerNorm, \textbf{h}=Linear. CIFAR dataset is CIFAR-10.}}
\end{table}

\textbf{Malicious Node Configuration:}  
The proportion of malicious nodes is fixed at 0\%, 10\%, 30\%, 40\% and 50\% on both the MNIST and CIFAR-10 datasets. The model accuracy is recorded over 50 training rounds to monitor its evolution. Malicious nodes are assumed to upload modified gradients dishonestly.

\textbf{Attack Types: }
Attacks are categorized into \textit{label flipping} (targeted) and \textit{gradient manipulation} (untargeted) attacks. Targeted attacks flip the label from $l$ to $N_l - l - 1$, where $N_l$ is the number of classes. Conversely, untargeted attacks replaces gradient updates with generated gaussian noise.

\subsection{Preliminary Experiment: }

\begin{figure}[htbp]
\centering
\label{LSH-Eucl}
\includegraphics[width=0.3\textwidth]{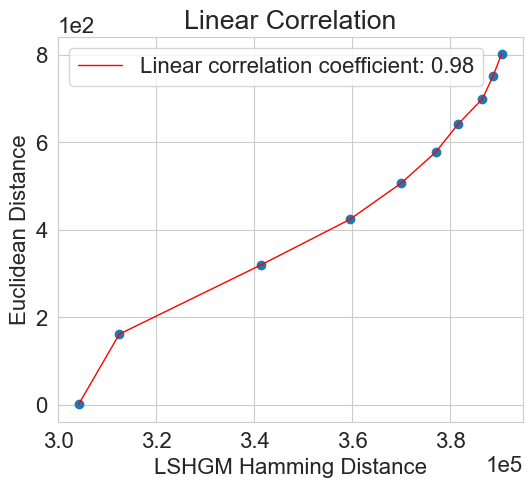}
\caption{\LSH-Euclidean Distance Correlation}
\label{LSH-Eucl}
\end{figure}

In the preliminary experiment, we explore the linear relationship between the number of differences produced by \LSH~and the Euclidean distance when calculating the same group of gradients. The experimental results demonstrates approximate linear correlation between the \LSH~distance and the Euclidean distance ($linear correlation > 0.98$), as shown in Figure \ref{LSH-Eucl}. The result means \LSH~provides detection performance comparable to that of Euclidean distance.

\subsection{Aggregation Performance} \label{Aggregation Performance}

\begin{table*}[ht]
\centering
\scriptsize
\resizebox{0.96\textwidth}{!}{%
\begin{tabular}{ll|ccccc|ccccc|c}
\toprule
\multirow{2}{*}{Dataset} & \multirow{2}{*}{Method} & \multicolumn{5}{c|}{Untargeted-attack} & \multicolumn{5}{c|}{Targeted-attack} & \multicolumn{1}{c}{Control group} \\
 & & 10\%& 20\%& 30\%& 40\%& 50\%& 10\%& 20\%& 30\%& 40\%& 50\%& - \\
\midrule

\multirow{5}{*}{MNIST (CNN)}
 & FedAvg& 0.89& 0.80& 0.71& 0.63& 0.53& 0.88& 0.78& 0.68& 0.59& 0.49& 0.98\\
 & FedATM \cite{nishimoto2023fedatm}& 0.63& 0.65& 0.76& 0.82& 0.82& 0.58& 0.60& 0.53& 0.45& 0.39& 0.98\\
 & CKADA \cite{ZHU2024110730}& 0.88& 0.78& 0.68& 0.67& 0.55& 0.88& 0.80& 0.70& 0.59& 0.55& 0.98\\
 & Krum \cite{colosimo2023median}& 0.98& 0.97& 0.98& 0.98& 0.83& 0.90& 0.75& 0.85& 0.57& 0.47& 0.98\\
 & Median \cite{xie2022robust}& 0.87& 0.84& 0.85& 0.78& 0.12& 0.80& 0.73& 0.65& 0.55& 0.44& 0.96\\
 & RFLPA \cite{mai2024rflpa}& 0.96& 0.95& 0.95& -& -& 0.96& 0.95& 0.95& -& -& 0.96\\
 \rowcolor{gray!20} 
 & \textbf{LSHFed (proposed)} & \textbf{0.98}& \textbf{0.98}& \textbf{0.97}& \textbf{0.97}& \textbf{0.97}& \textbf{0.97}& \textbf{0.97}& \textbf{0.97}& \textbf{0.97}& \textbf{0.97}& \textbf{0.98}\\

\midrule
\multirow{5}{*}{CIFAR-10 (ResNet)}
 & FedAvg& 0.91& 0.82& 0.73& 0.64& 0.50& 0.90& 0.80& 0.70& 0.60& 0.50& 0.99\\
 & FedATM \cite{nishimoto2023fedatm}& 0.12& 0.12& 0.17& 0.14& 0.18& 0.10& 0.11& 0.10& 0.18& 0.17& 0.99\\
 & CKADA \cite{ZHU2024110730}& 0.89& 0.83& 0.77& 0.64& 0.47& 0.90& 0.81& 0.70& 0.58& 0.53& 0.99\\
 & Krum \cite{colosimo2023median}& 0.99& 0.99& 0.99& 0.99& 0.58& 0.85& 0.85& 0.73& 0.63& 0.48& 0.99\\
 & Median \cite{xie2022robust}& 0.58& 0.54& 0.51& 0.46& 0.22& 0.64& 0.59& 0.51& 0.44& 0.38& 0.71\\
 & RFLPA \cite{mai2024rflpa}& 0.70& 0.70& 0.69& -& -& 0.71& 0.70& 0.69& -& -& 0.74\\
 \rowcolor{gray!20} 
 & \textbf{LSHFed (proposed)} & \textbf{0.99}& \textbf{0.99}& \textbf{0.99}& \textbf{0.99}& \textbf{0.99}& \textbf{0.99}& \textbf{0.99}& \textbf{0.99}& \textbf{0.99}& \textbf{0.99}& \textbf{0.99}\\

 \midrule
 \multirow{5}{*}{F-MNIST (ViT)}
 & FedAvg& 0.77& 0.69& 0.62& 0.55& 0.47& 0.88& 0.80& 0.70& 0.59& 0.50& 0.88\\
 & FedATM \cite{nishimoto2023fedatm}& 0.73& 0.67& 0.60& 0.51& 0.41& 0.72& 0.64& 0.56& 0.48& 0.40& 0.88\\
 & CKADA \cite{ZHU2024110730}& 0.76& 0.67& 0.62& 0.56& 0.41& 0.78& 0.70& 0.54& 0.55& 0.46& 0.88\\
 & Krum \cite{colosimo2023median}& 0.85& 0.85& 0.85& 0.85& 0.65& 0.72& 0.76& 0.64& 0.57& 0.41& 0.87\\
 & Median \cite{xie2022robust}& 0.62& 0.57& 0.52& 0.47& 0.41& 0.61& 0.54& 0.47& 0.40& 0.34& 0.75\\
 \rowcolor{gray!20}
 & \textbf{LSHFed (proposed)} & \textbf{0.87}& \textbf{0.88}& \textbf{0.87}& \textbf{0.87}& \textbf{0.88}& \textbf{0.88}& \textbf{0.88}& \textbf{0.87}& \textbf{0.88}& \textbf{0.87}& \textbf{0.88}\\

\bottomrule
\end{tabular}%
}
\caption{Results of Aggregation Performance against Malicious Attacks}
\footnotesize{\textit{* The control group refers to the scenario where all participating nodes are honest, i.e., no adversarial behavior is introduced during training. This serves as an upper-bound reference for evaluating the robustness of each method under attack.}}
\label{generalanal}
\end{table*}

\begin{figure*}[t]
  \centering
  \includegraphics[width=0.8\textwidth]{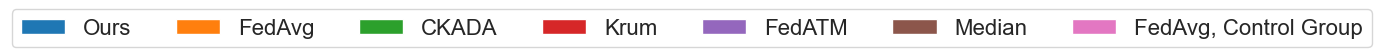}
  \vspace{1ex}

  \begin{tabular}{ccc}
    \includegraphics[width=0.21\textwidth]{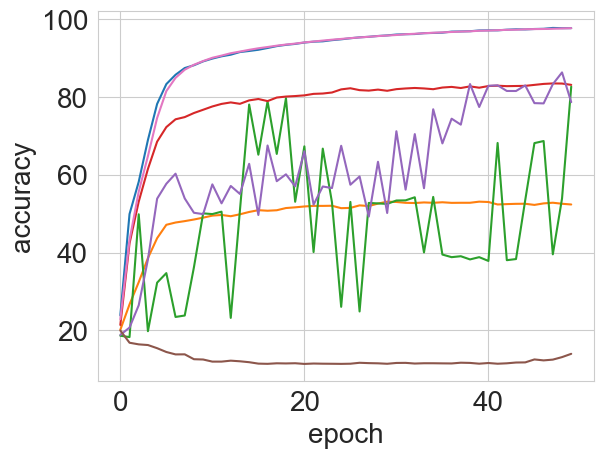} &
    \includegraphics[width=0.21\textwidth]{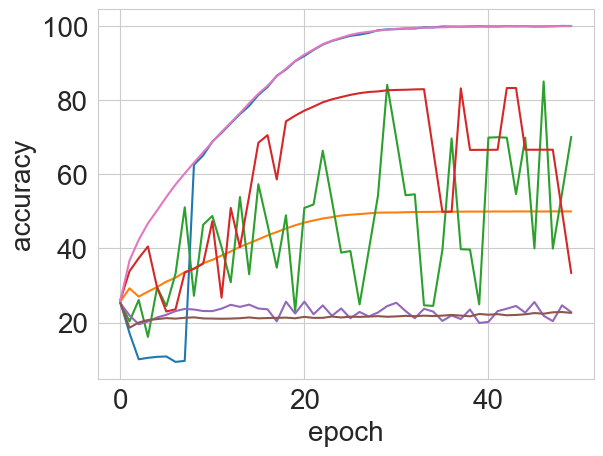} &
    \includegraphics[width=0.21\textwidth]{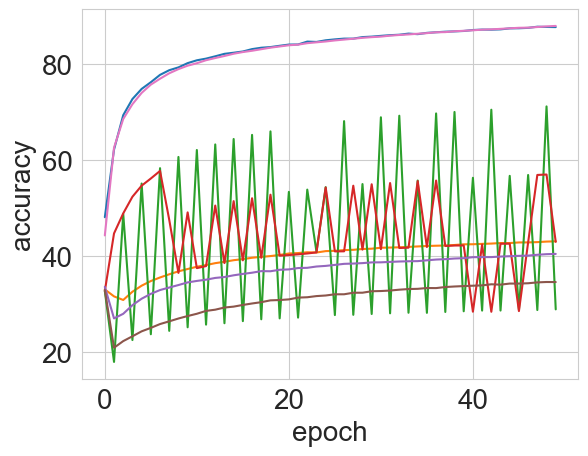} \\[2pt]
    (a) MNIST 50\% malicious &
    (b) CIFAR-10 50\% malicious &
    (c) FMNIST 50\% malicious
  \end{tabular}

  \caption{Per-iteration Model Performance}
  \label{feimubiao21}
\end{figure*}

To evaluate the overall framework's resistance to malicious attacks, robustness, and aggregation model accuracy, experiments are conducted upon \OurFed. In targeted attacks, all malicious nodes act in collusion; in untargeted attacks, they operate independently. 

Table \ref{generalanal} exhibits the aggregation performance of LSHFed compared to a series of baseline methods under varying proportions of malicious clients. The final accuracy is measured as the average accuracy over the last 10 training epochs. The baselines include RFLPA \cite{mai2024rflpa}, FedAvg, FedATM \cite{nishimoto2023fedatm}, CKADA \cite{ZHU2024110730}, Krum \cite{colosimo2023median}, and Median \cite{xie2022robust}. Notably, RFLPA only supports up to 30\% malicious clients and was not evaluated on the ViT–based F-MNIST setting; therefore, its results are unavailable for the 40\% and 50\% and F-MNIST settings.

As shown in Table~\ref{generalanal}, LSHFed consistently achieves the highest overall accuracy across all datasets and attack scenarios, and even when 50\% of the clients are adversarial it maintains performance on par with the benign (0\%) condition. This resilience stems from LSHFed’s ability to selectively aggregate only non-malicious updates, thereby preserving near–no–attack accuracy until the system’s theoretical failure point. In contrast, while Krum also demonstrates relatively strong accuracy at lower adversary fractions, it tends to collapse once the proportion of malicious participants exceeds 40\%. 

\begin{table*}[ht]
\centering
\resizebox{0.96\textwidth}{!}{%
\begin{tabular}{l|c|c|c|c|c}
\toprule
Method & Client computation (LT \& AG) & Communication per client  & Server computation (VR) & Server Comm. (first \& last round) & Server Comm. (other rounds) \\
\midrule
RFLPA  & \(\mathcal{O}((R^2+N)\log^{2}R)\) & \(\mathcal{O}(R+N)\) & \(\mathcal{O}((R+N)\log^{2}R\log\log R)\) & \(\mathcal{O}((R+N)\,R)\) & \(\mathcal{O}((R+N)\,R)\) \\[6pt]
\OurFed  & \(\mathcal{O}\bigl(R_{AG}\,\sum_{i=1}^{N}n_i\,r\bigr)\;\approx\;\mathcal{O}(R)\) 
         & \(\mathcal{O}(M + \sum_{i=1}^{N}n_i\,r)\;\approx\;\mathcal{O}(R+N)\) 
         & \(\mathcal{O}\bigl(R_{AG}\,\sum_{i=1}^{N}n_i\,r\bigr)\;\approx\;\mathcal{O}(R)\) 
         & \(\mathcal{O}(R \times M)\;\approx\;\mathcal{O}(R N)\) 
         & \(\mathcal{O}\bigl(R_{AG}\,\sum_{i=1}^{N}n_i\,r\bigr)\;\approx\;\mathcal{O}(R)\) \\
\bottomrule
\end{tabular}%
}
\caption{Communication and Computation Comparison between RFLPA and \OurFed}
\label{overhead}
\end{table*}

\subsection{Non-IID Robustness}
We also evaluate the performance of LSHFed in non-iid settings, as presented in Table~\ref{noniid}. Two types of data heterogeneity are considered: \textit{label skew} and \textit{Dirichlet} distributions. The experiments are conducted on the F-MNIST dataset using a ViT model. The model configuration is consistent with that listed in Table~\ref{tab:exp-config}.
In the label skew scenario, each client receives 6000 training samples, where 70\% of the labels are drawn from two common primary classes shared across clients, and the remaining 30\% are randomly sampled from the other eight classes. In the Dirichlet setting, the training data is allocated among clients based on a Dirichlet distribution, which models varying degrees of label distribution skewness. Specifically, for each class, samples are distributed across three clients using Dirichlet concentration parameter $\alpha = 0.5$.

\begin{table}[ht]
\centering
\resizebox{0.48\textwidth}{!}{%
\begin{tabular}{l|ccccc|c}
\toprule
\multirow{2}{*}{Data distribution}  & \multicolumn{5}{c|}{Targeted-attack} & \multicolumn{1}{c}{Control group} \\
 & 10\% & 20\% & 30\% & 40\% & 50\% & - \\
\midrule
Dirichlet & 0.91& 0.91& 0.91& 0.91& 0.91& 0.91\\
\midrule
Label skew partition & 0.88& 0.88& 0.87& 0.87& 0.87& 0.88\\
\bottomrule
\end{tabular}%
}
\caption{LSHFed in Non-IID Scenario}
\label{noniid}
\end{table}

\subsection{Scalability}
To assess the scalability of LSHFed, we further evaluate its performance under varying numbers of participating clients. The results are summarized in Table~\ref{100node}. The experiments are conducted on the F-MNIST dataset using a ViT model. The model configuration is consistent with that listed in Table~\ref{tab:exp-config}.

\begin{table}[ht]
\centering
\resizebox{0.48\textwidth}{!}{%
\begin{tabular}{l|c|c}
\toprule
    Node Count & Accuracy in 50\% Targeted-attack Scenario& Control Group \\
    \midrule
    10 Nodes  & 0.88 & 0.88 \\
    20 Nodes  & 0.88 & 0.88 \\
    100 Nodes & 0.85 & 0.88 \\
    \bottomrule
\end{tabular}%
}
\caption{LSHFed Scalability}
\label{100node}
\end{table}

\subsection{Communication \& Computation Overhead}

In this section, we provide both theoretical evaluations of the communication and computation overhead of LSHFed. Table~\ref{overhead} presents a theoretical comparison between LSHFed and RFLPA in terms of per-round communication and computation costs. Assume each client’s gradient update comprises \(N\) matrices and \(M\) parameters, and its LSH bit‐string has length proportional to \(\sum_{i=1}^{N} n_i \times r\) (with \(r\) hyperplanes). Let \(R\) be the total number of nodes, decomposed as \(R_{LT}\) local trainers and \(R_{AG}\) aggregators, and \(R_{VR}=1\) verifier.

\section{Conclusion}
We introduce a novel FL framework, \OurFed, which aims at improving the robustness of the aggregation process at a low cost. \OurFed~enhances the aggregation process by incorporating a mask-based privacy protection scheme and the Score-Q Hash role election algorithm, effectively detecting and excluding malicious nodes while protecting gradient privacy. The core module in \OurFed, namely \LSH, employs a multi-hyperplane projection technique to irreversibly map high-dimensional gradients into compact binary bit strings. By measuring the Hamming distance of the bit strings, \LSH~offers a communication- and computation-efficient method for identifying malicious gradients. Experimental results confirm that \OurFed~outperforms baseline robust FL frameworks, maintaining high model accuracy even when up to 50\% of participating nodes are malicious and reducing the communication overhead of gradient verification by at most three orders of magnitude than using full gradients.

\bibliography{aaai2026}

\end{document}